\documentclass[12pt]{article}
\usepackage{graphicx}
\usepackage{multirow}
\usepackage{booktabs}
\usepackage{amsmath}
\usepackage[a4paper, total={6in, 9in}]{geometry}
\usepackage[font=footnotesize]{caption}
\usepackage{hyperref}
\usepackage{lineno}
\usepackage{makecell}
\usepackage[ruled]{algorithm2e}
\usepackage{algpseudocode}
\begin{document}
\title{Continual Learning of Multiple Cognitive Functions with Brain-inspired Temporal Development Mechanism}

\author
{Bing Han$^{1,3,\#}$, Feifei Zhao$^{1,\#}$, Yinqian Sun$^{1}$, Wenxuan Pan$^{1,3}$, Yi Zeng$^{1,2,3,4,*}$\\
\\
\normalsize{$^{1}$Brain-inspired Cognitive Intelligence Lab, }\\
\normalsize{Institute of Automation,Chinese Academy of Sciences}\\
\normalsize{$^{2}$State Key Laboratory of Brain Cognition and Brain-inspired Intelligence Technology, }\\
\normalsize{Chinese Academy of Sciences}\\
\normalsize{$^{3}$ School of Artificial Intelligence, University of Chinese Academy of Sciences}\\
\normalsize{$^{4}$ Center for Long-term Artificial Intelligence}\\
\normalsize{$^{*}$Corresponding authors: yi.zeng@ia.ac.cn}\\
\normalsize{$^{\#}$Co-first authors with equal contribution}\\
}

\maketitle


\begin{abstract}
Cognitive functions in current artificial intelligence networks are tied to the exponential increase in network scale, whereas the human brain can continuously learn hundreds of cognitive functions with remarkably low energy consumption. This advantage is in part due to the brain’s cross-regional temporal development mechanisms, where the progressive formation, reorganization, and pruning of connections from basic to advanced regions, facilitate knowledge transfer and prevent network redundancy. Inspired by these, we propose the Continual Learning of Multiple Cognitive Functions with Brain-inspired Temporal Development Mechanism(TD-MCL), enabling cognitive enhancement from simple to complex in Perception-Motor-Interaction(PMI) multiple cognitive task scenarios. The TD-MCL model proposes the sequential evolution of long-range connections between different cognitive modules to promote positive knowledge transfer, while using feedback-guided local connection inhibition and pruning to effectively eliminate redundancies in previous tasks, reducing energy consumption while preserving acquired knowledge. Experiments show that the proposed method can achieve continual learning capabilities while reducing network scale, without introducing regularization, replay, or freezing strategies, and achieving superior accuracy on new tasks compared to direct learning. The proposed method shows that the brain's developmental mechanisms offer a valuable reference for exploring biologically plausible, low-energy enhancements of general cognitive abilities.

\end{abstract}

\section*{Keywords}
 Brain-inspired Temporal Development, Multiple Cognitive Functions Continual Learning, Evolutionary Growth Long-range Connectivity,  Feedback-guided Suppression and Pruning,  Biological Synaptic Plasticity

\section{Introduction}
Artificial intelligence algorithms have achieved remarkable success across various fields, but their enhancement of cognitive functions relies on the massive stacking of parameters, often facing challenges in balancing memory capacity with energy consumption\cite{wickramasinghe2023continual}. In contrast, the brain requires only 20 watts of power to gradually master a rich array of cognitive functions during its developmental process, offering valuable biological insights. Biological research shows that the development of brain regions follows a specific chronological sequence: from primary to higher brain regions\cite{huttenlocher1997regional}, from intra-regional to inter-regional connections\cite{johnson2001functional}, from active growth to inhibitory pruning~\cite{huttenlocher1979synaptic}, occurring at different ages in children that progressively establish well-rounded cognitive functions. Notably, advanced cognitive functions evolve from foundational primary functions\cite{flavell1982cognitive}, and this progressive development allows the brain to maintain its capacity to continually learn new cognitive tasks even during synaptic pruning and neuronal reduction.

To enable artificial neural networks to learn multiple tasks efficiently, extensive research has been conducted in the fields of continual learning, multi-task learning, and transfer learning. Specifically, multi-task learning \cite{zhang2021survey,le2024continual,marza2024task} primarily addresses weight update conflicts among concurrent tasks but overlooks the temporal dependencies between tasks. Transfer learning\cite{zhuang2020comprehensive,he2024efficient,zhao2024comparison} leverages prior knowledge to facilitate new task learning, yet struggles to prevent catastrophic forgetting of previously learned tasks. While continual learning \cite{van2019three,wang2023task,han2025similarity}mitigates catastrophic forgetting and enables progressive learning of multiple tasks within the same domain, it remains constrained by limitations such as task homogeneity and inefficient knowledge transfer. More critically, existing approaches often incur substantial computational costs or performance degradation when expanding memory capacity \cite{hu2023dense}. It remains an important challenge to realize cross-domain multi-task progressive continual learning with brain-like adaptivity and low-power consumption.

The brain development process follows the basic characteristics: 1) Overall connectivity increases first and then decreases\cite{huttenlocher2009neural}, in which inter-regional long-range connectivity gradually increases\cite{hagmann2010white}, and local connectivity first grows explosively and then gradually prunes\cite{huttenlocher1979synaptic}.  2) Primary brain regions develop earlier than higher brain regions\cite{huttenlocher1997regional}, and the pruning rate is greater than that of higher brain regions\cite{selemon2013role}. 3) Primary cognitive brain regions support the learning of complex higher cognitive functions\cite{flavell1982cognitive}, and higher cognitive feedback guides structural optimization in primary brain regions\cite{scott2004optimal,sitaram2017closed}.  Meanwhile, the learning of multiple cognitive functions in infants does not happen overnight, but follows a progression from simple visual-tactile perception, to motor control of the body, to complex cognitive functions such as reasoning and decision-making through interaction with others or objects\cite{flavell1982cognitive}. The multi-scale developmental rules of the infant brain at multiple developmental time stages are manifested in the individual's experiential learning as adaptive regulation of the growth and extinction of connections\cite{finlay2001developmental}, and ultimately the brain realizes continual learning of multiple cognitive functions from simple to complex with very low energy consumption.


Existing brain development-inspired artificial neural network algorithms primarily focus on structural compression. They employ fine-grained\cite{shi2023towards} or structured compression \cite{yu2022width} \cite{balaskas2024hardware} based on parameters such as network weights\cite{molchanov2016pruning}, BN factors\cite{you2019gate}, and similarity metrics\cite{srinivas2015data}, aiming to minimize energy consumption. In particular, brain-inspired Spiking Neural Networks (SNNs) provide an excellent basic platform for brain mechanism simulation with efficient and highly bio-interpretable spiking delivery~\cite{Maass1997Networks,gerstner2002spiking}. Current brain-inspired structural optimization algorithms for SNNs can be classified into: synaptic plasticity pruning\cite{rathi2018stdp},\cite{han2024developmental}, neural activity pruning\cite{wu2019adaptive},\cite{liu2022dynsnn}, and pruning-growth fusion algorithms\cite{chen2022state,han2025adaptive}. These methods reduce network energy consumption while following biological pruning and growth principles. However, they only focus on energy reduction through isolated growth-pruning mechanisms, lacking integrated cross-regional temporal development to continual learning of novel complex tasks and forward knowledge transfer during network scaling-down.

To  support low-energy multiple cognitive task continual learning, we propose the Continual Learning of Multiple Cognitive Functions with Brain-inspired Temporal Development Mechanism (TD-MCL) of the children's brain. The TD-MCL algorithm models the inter-regional temporal developmental process of the brain realizing progressive learning of perceptual classification - body control - environmental interaction multiple cognitive functions. Specifically, following the developmental principles of brain regions (from primary to advanced) and the hierarchical organization of cognitive tasks (from simple to complex), TD-MCL uses evolutionary algorithms to progressively build SNN modules and strengthen cross-region long-range connections starting from basic perceptual functions. Adaptive learning of interaction patterns between old and new tasks enables old tasks to facilitate learning of new tasks as much as possible. Correspondingly, as learning progresses, subsequent complex task feedback guides the transition of local connectivity within the earlier task module from an active to an inhibited state. Numerous local connections from early tasks that are not activated in new tasks gradually weaken or disappear, leading to a reduction in overall network size without interrupting continual task learning. We demonstrate in SNN that, without introducing continual learning training regularization, sample replay, or parameter freezing, TD-MCL achieves the capability of continual learning from simple to complex tasks in progressively scaled-down networks and enhances the learning performance of new cognitive tasks.

\section{Results}

\subsection{multiple cognitive function Dataset}

\begin{figure*}[t]
	\centering 
	\includegraphics[width=0.98\linewidth]{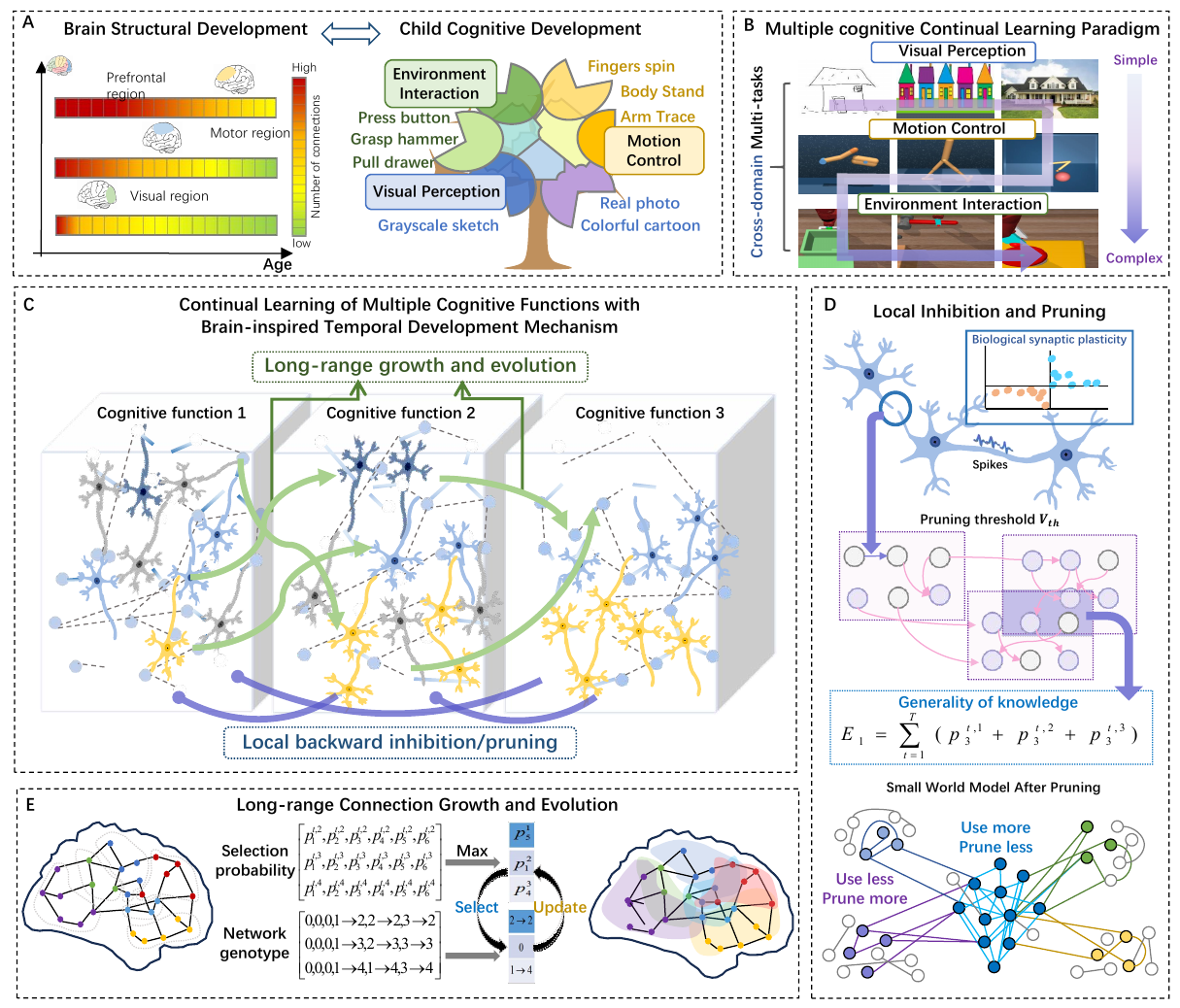}
	\caption{\textbf{The procedure of TD-MCL model.}  \textbf{A)} Correspondence between cognitive function development and brain structure development in children. \textbf{B)} Multiple Cognitive Functions Continual Learning Dataset Paradigm. \textbf{C)} General overview of the procedure of efficient continual multiple cognitive function learning algorithm. \textbf{D)}  Local connection suppression and pruning integrating biological synaptic plasticity and knowledge generalization. \textbf{E)}  Long-range connection growth based on evolutionary algorithms. }
	\label{f61}
\end{figure*}

The development of children's cognitive functions exhibits temporal characteristics, progressing from basic to advanced in a hierarchical sequence during specific sensitive periods. In the first two months of birth, infants explore the world primarily through sensory input (visual, auditory, and tactile); between 2-10 months, they begin to combine perception with movement to perform simple limb movements; and at about 1 year of age, hand-eye coordination improves and child can grasp objects and interact with the outside world \cite{inguaggiato2017brain}. Among them, the development of higher cognitive functions is based on primary ones (as Fig. \ref{f61}A, right). However, existing continual learning methods focus only on a single cognitive function, such as visual perception classification or motor control, while lacking exploration of cross-domain continual learning.

To model the progressive of children cognitive functions, we designed a \textbf{Perception-Motor-Interaction} (PMI) \textbf{cross-domain} dataset (as Fig. \ref{f61}B), which incorporates nine tasks spanning three cognitive functions and arranges in a sequentially progressive order from simple to complex. Specifically, the first perceptual domain consists of three visual recognition tasks\cite{zhou2020deep}: sketch, cartoon, and photo. The motor domain consists of waving an arm to reach a specified position, standing up, and finger-toggling limb movement tasks\cite{tassa2018deepmind}. The interaction tasks include pressing the button, pulling the drawer and grasping the hammer to knock the object tasks interacting with the external world\cite{yu2020meta}. The deep SNN progressively learn the above nine tasks during training, and is evaluated on all learned tasks during testing.

\subsection{Efficient continual multiple cognitive function learning algorithm}

Cognitive function development parallels brain structural development as Fig. \ref{f61}A. Inspired by the brain multi-scale temporal development, we propose the efficient continual multiple cognitive function learning algorithm incorporating temporal progressive module growth, long-range connectivity growth and evolution, and local connectivity inhibition and pruning as Fig. \ref{f61}C.

\textbf{Temporal progressive module growth.} Neuroscientific researches have shown that overall brain synaptogenesis increases rapidly around birth\cite{huttenlocher2013synaptogenesis,bourgeois1993changes}, but the onset and peak time of synaptic bursts in different regions occur at different ages\cite{huttenlocher1997regional}. For example, synapse formation in the visual cortex bursts rapidly between 3-12 months, reaching maximum densities about 150\% of adults; bursts of the prefrontal cortex only begin at this time and do not peak until one age\cite{kroon2019early}. Thus, the SNN modules of our network are not fixed at the beginning of learning, but grow progressively following the learning sequence of perceptual-motor-interactive tasks, and connect them to form a unified network.

\textbf{Long-range connectivity growth and evolution.} During subsequent development, inter-regional long-range communication continued to increase\cite{wilke2007global,groeschel2010developmental,giedd2015child}, which facilitated the efficient reuse of the acquired repetitive knowledge part of each task \cite{reveley2015superficial,schuz2002human}. Therefore, we employed the \textbf{online evolutionary algorithm} to adaptively enhance long-range connectivity by \textbf{selecting beneficial modules} from previous tasks to support new ones and promote positive knowledge transfer. The evolution space of a new task  module (excluding input and output modules) consists of possible connections to all learned task  modules and no connections established. During update, we increase the connection probability with high historical performance and low number of choices as Fig. \ref{f61}E.

\textbf{Local connectivity inhibition and pruning.} 
Low-level cognitive brain regions lay the foundation for higher functions, and the development of higher cognition subsequently optimizes the sparse structure of low-level brain regions\cite{scott2004optimal,sitaram2017closed}. Large numbers of localized dendritic spines and synapses that are not useful in higher functions are progressively suppressed or even pruned in developmental temporal order\cite{huttenlocher1979synaptic,chechik1999neuronal,huttenlocher2013synaptogenesis}, reducing the response to irrelevant stimuli and improving attentional control\cite{stephan2012complement,faust2021mechanisms,ugarte2023attention}. For example, pruning begins earlier in the auditory cortex and ends at age 12, while pruning in the prefrontal cortex extends through mid-puberty\cite{huttenlocher1997regional}. Inspired by this, we provided feedback during the learning of complex tasks to guide local connectivity inhibition and pruning for already learned simple tasks as Fig. \ref{f61}D. The evaluation of connection importance combines \textbf{local synaptic plasticity} with \textbf{global knowledge generality}, where the former is the integration of pre- and post-synaptic neuron activations, and the latter is the sum of the evolutionary selection probabilities of the SNN module in subsequent tasks. This \textbf{feedback-guided inhibition and pruning} enable precise identification and retention of globally-utilized connections while pruning redundant ones from learned tasks, thereby achieving continual learning without employing regularization, replay, or freezing.

\subsection{Efficient Continual Learning Performance Improvement}
\begin{figure*}[t]
	\centering 
	\includegraphics[width=0.98\linewidth]{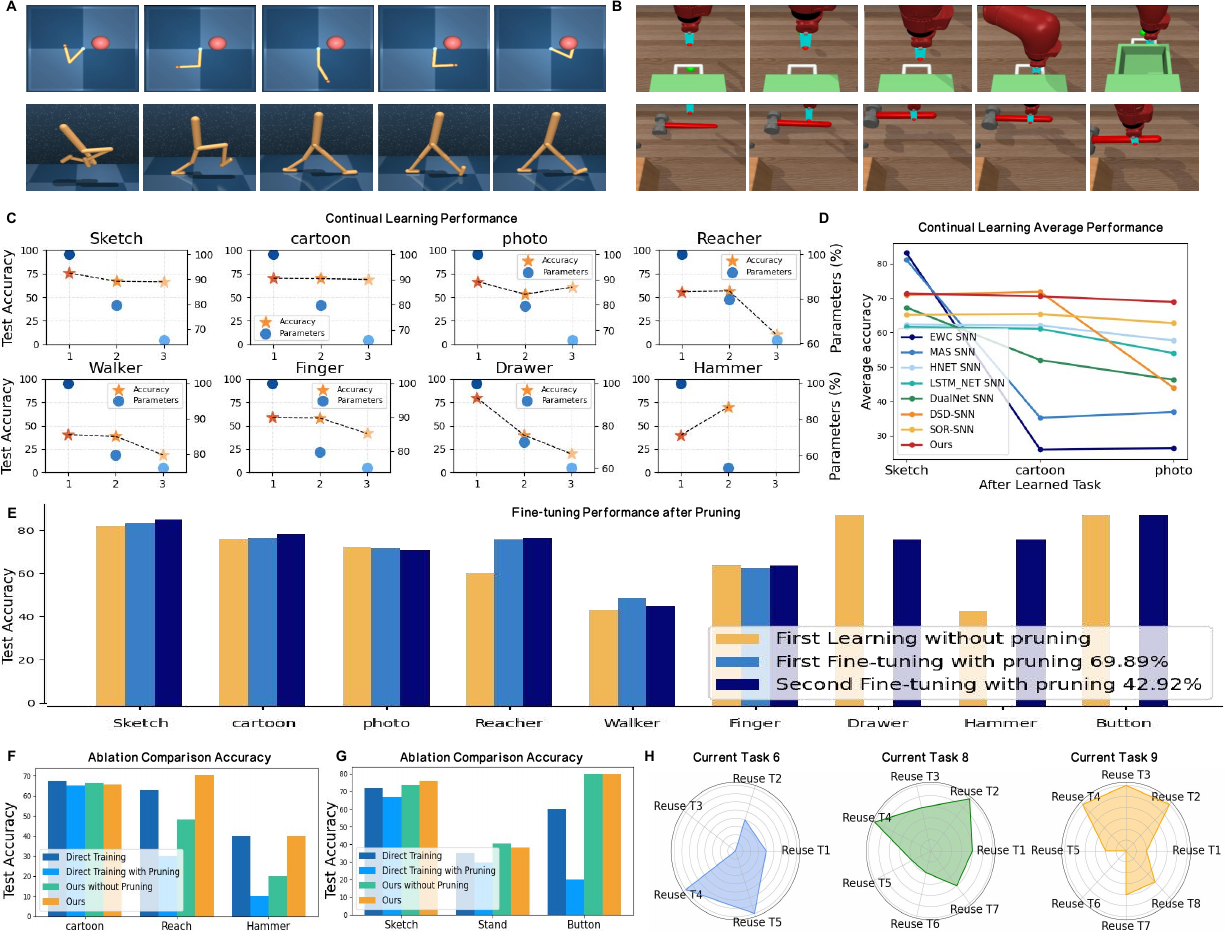}
	\caption{\textbf{Progressive continual learning performance.} \textbf{A-B)} Examples of successful processes for motion control and environmental interaction tasks. \textbf{C)} Effect of pruning on continual learning performance. \textbf{D)} Fine-tuning performance of sparse networks after pruning. \textbf{E)} Comparison with other continual learning methods. \textbf{F-G)} Performance of ablation experiments. \textbf{H)} Module reuse determined by long-range connectivity.}
	\label{f62}
\end{figure*}

Fig. \ref{f62}C presents the corresponding changes in network parameter quantity and performance for the first eight tasks after feedback pruning (with Task 8 undergoing only a single pruning guided by the ninth task), without incorporating any replay, regularization, or freezing operations typically used in continual learning. The results demonstrate that, after feedback-inhibitory pruning, the performance of previously learned tasks is well preserved as the network size for each task exponentially decreases. For instance, the performance of tasks 1, 2, 3, and 6 remain essentially unchanged, at which point their number of parameters is 60.66\%, 60.99\%, 61.09\% and 77.53\% respectively. Instead of forgetting occurring after pruning 46.89\% Notably, Task 4 exhibits a 0.84\% performance improvement while maintaining 79.81\% of its parameters (The successful task of motion control is shown in Fig. \ref{f62}A). Similarly, Task 8 achieves a 70.00\% success rate despite a significant 46.89\% reduction in redundant network parameters(The successful task of environment interaction as Fig. \ref{f62}B). Although Tasks 4, 5, and 7 experience slight forgetting, for example, Task 5 shows only a marginal performance decline of 1.93\% at 79.73\% parameter retention—they still retain learning capability. These findings indicate that the proposed algorithm successfully maintains memory of previously learned tasks while acquiring new ones, even as the network size  exponentially decreases and sparsity increases. The proposed algorithm successfully acquires new knowledge while preserving old ones as the network becomes more compact and sparse, \textbf{without relying on conventional continual learning constraints.}

To demonstrate the effectiveness of our approach, we conducted comparative experiments with existing continual learning algorithms. Due to our algorithm providing a unified continual learning framework for perception-motion-interaction across cognitive functions, while most current algorithms focus on single cognitive functions, we conducted a comparative analysis in the widely studied domain of visual perception, as shown in Fig. \ref{f62}D, including: the DNN-based continual learning algorithms EWC~\cite{kirkpatrick2017overcoming}, MAS~\cite{aljundi2018memory}, HNET~\cite{von2019continual},  LSTM$\_$NET~\cite{chandra2023continual}, and the DualNet~\cite{pham2021dualnet} migrated to SNN, and the SNN-based continual learning algorithms DSD-SNN~\cite{han2023enhancing}, and SOR-SNN\cite{han2023adaptive}. 

Our algorithm achieved the highest average accuracy of 68.88\% on learned tasks, representing a 6.15\% improvement over the second-best performer SOR-SNN. Although our algorithm did not attain peak performance in the first simple task, it demonstrated remarkable late-stage acceleration, exhibiting superior capability in learning complex tasks compared to other methods. For instance, in the photo task, our algorithm achieved 66.00\% accuracy, outperforming MAS SNN, HNET SNN, and DSD-SNN by 12.67\%, 20.00\%, and 3.54\% respectively. Notably, while LSTM$\_$NET SNN, DSD-SNN, EWC SNN and MAS SNN all exhibited significant performance degradation throughout the learning process, our algorithm maintained the most stable continual learning performance. These experimental results confirm that the proposed algorithm has reached superior performance comparable to existing single-domain continual learning algorithms.

Further confirming the superiority of the sparse architecture after feedback pruning, we performed fine-tuning on the pruned models of previous tasks when the total pruning rate reached 69.89\% (after the 5th task) and 42.92\% (after the 9th task), as shown in Fig. \ref{f62}E. For example, the initial learning accuracy of Task 1 (sketch recognition) was 75.64\%, and after feedback inhibition across eight subsequent tasks, the pruning rate of the Task 1 module reached 53.79\%, yet fine-tuning led to an accuracy improvement to 78.22\%. More notably, the initial accuracy of Task 4 (arm tracking) was only 45.28\%, but after pruning 49.62\% of the connections, the accuracy during the first fine-tuning improved significantly to 70.47\%. Results show that our algorithm \textbf{precisely removes redundant and useless connections} through feedback pruning for new tasks, helping preserve key synapses from old tasks. As a result, it not only prevents forgetting but also achieves significant improvement with minimal fine-tuning.

\subsection{Cognition Gradually Progresses from Simple to Complex}
To demonstrate the superiority of the progressive learning mode from simple to complex for the proposed algorithm, we compare it with the direct training and direct pruning optimization methods (Independent learning of individual tasks) for the same network, as shown in Fig. \ref{f62}F-G. The experimental results show that our algorithm achieves the highest performance for the same network size and the same pruning rate. For example, in the arm motion task, the accuracy of direct training and direct pruning is only 29.9\% and 63.02\%, while our algorithm significantly improves the performance to 70.33\% based on cartoon drawing learning. In the button-pressing task, direct training yielded only 60\% accuracy, which dropped sharply to 20\% after 40\% parameter pruning. By contrast, our algorithm achieved 80\% accuracy despite a higher 51.46\% pruning rate based on prior stick-figure and standing tasks. These findings highlight the positive role of the progressive learning mode, where \textbf{new tasks build upon previously learned ones}, in enhancing the performance of neural networks.

The above improvement is attributed to the the feedback inhibition mechanism that prunes redundant individual and irrelevant knowledge, as well as the effective identification and reuse of shared visual knowledge through long-range connections. Therefore, we compare the performance of networks with long-range evolution but without local inhibition(as Fig. \ref{f62}F-G). The results show that the performance is lower than or approximately equal to the full TD-MCL algorithm. This indicates that the proposed algorithm achieves a significant reduction in network size without a substantial decline in performance. Additionally, we quantified the evolutionary selection count of long-range connections across nine tasks in Fig. \ref{f62}H. The results show that task 6 (finger-related) mainly connects with congruent cognitive Task 4 (arm) and Task 5 (standing), followed by visually similar tasks like sketches and cartoons, while avoiding unrelated real photos. Also, task 9 (button pressing) connects primarily with cognate cognitive task 7 (drawer pulling) and task 8 (hammering), along with related arm movement and color-based tasks.

 \subsection{Dynamic network connections promote knowledge transfer and reduce energy consumption}

\begin{figure*}[t]
	\centering 
	\includegraphics[width=0.98\linewidth]{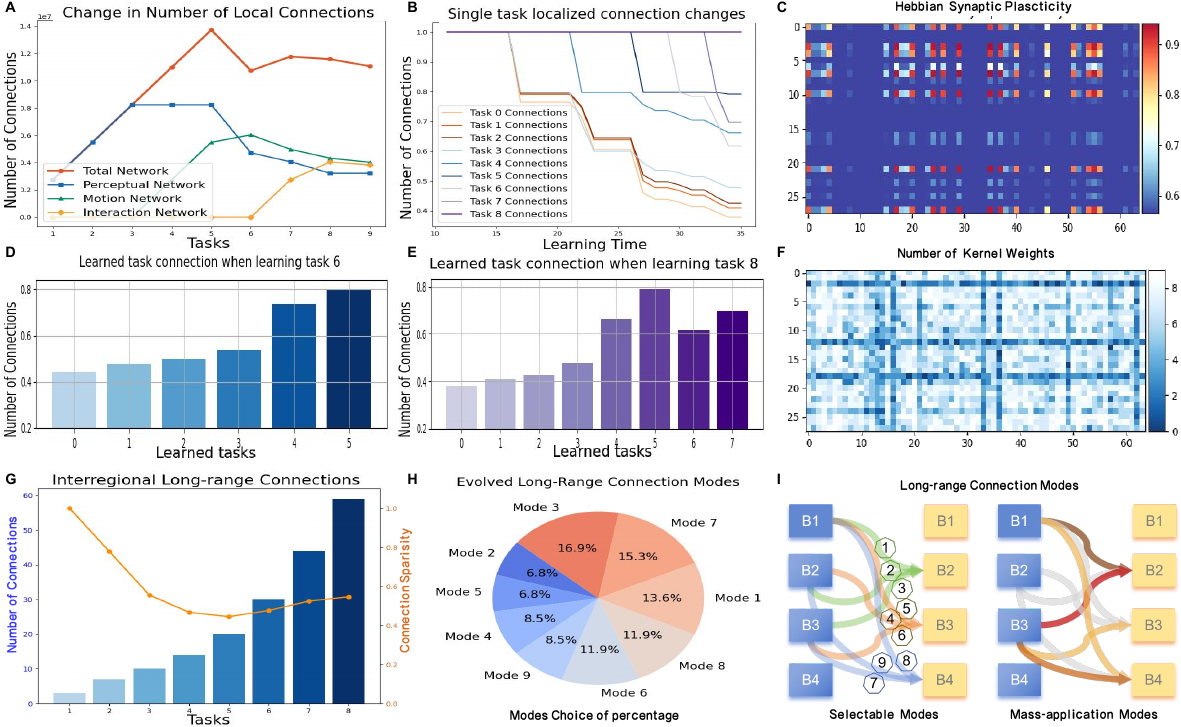}
	\caption{\textbf{Local connectivity and long-range connectivity dynamics.} \textbf{A)} Local connectivity changes in the overall and cognitive-functional networks.\textbf{B)} Local connectivity changes in single-task networks. \textbf{D-E)} Number of retained parameters from previous tasks. \textbf{C, F)} Correlation of biological synaptic plasticity with local connection pruning.  \textbf{G)} Long-range connection parameter counts and sparsity. \textbf{H-I)} Long-range connection mode selection.}
	\label{f63}
\end{figure*}

To understand the dynamics of feedback pruning and knowledge retention, Fig. \ref{f63}A illustrates the changes in local connections under feedback inhibition and pruning across the overall network and three cognitive functions: perception, motor control, and interaction. The results show that the total number of local connections first increases, then decreases, and finally stabilizes. This pattern aligns with the biological brain, which grows rapidly before age 2 and prunes between ages 2 and 10 while maintaining the ability to learn new tasks continuously\cite{huttenlocher1979synaptic}. At the cognitive function level (Fig. \ref{f63}A, fine line) and the individual task level (Fig. \ref{f63}B), a similar trend appears. Simple perception modules grow and prune earlier with a higher pruning rate, while complex interaction modules learn later and retain more connections\cite{selemon2013role}. For example, pruning in the visual module mainly occurs during tasks 5-7, with the final number of parameters reduced to approximately 39.14\% of the peak value; whereas pruning in the motor module is concentrated between tasks 6-8, with the parameters after pruning representing 66.70\% of the peak value.  Additionally, we examined the relationship between biologically inspired synaptic plasticity and pruning rates at the convolutional kernel level in Fig. \ref{f63}F. The results show that the algorithm adaptively prunes more synapses in convolutional kernels with lower synaptic plasticity. These results show that the proposed algorithm's cross-region network growth and pruning \textbf{follow the biological brain's temporal developmental patterns}\cite{huttenlocher1997regional}, with feedback-guided pruning removing redundant connections and enhancing attention to improve task performance\cite{faust2021mechanisms}.

Furthermore, from the perspective of forward knowledge transfer, we monitored the post-pruning network size of previous tasks used by the current task  (Fig. \ref{f63}D-E). For example, for motor control task 6, the pruning rate of the adjacent learned task 5 is the smallest, while the pruning rate of the initial task 1 is the largest. This suggests that the influence of adjacent learned tasks on the current task is the greatest, and the knowledge of previous tasks progressively transfers backward through the incremental learning process\cite{favazza2016motor}, enabling large-scale pruning of earlier tasks without significant performance degradation.

 For long-range connectivity, Fig. \ref{f63}G shows the changes in the number and sparsity of long-range connections during progressive learning. In the progressive learning process, as the number of available task modules for the current task gradually increases, the number of long-range connections gradually increases, in accordance with the temporal developmental patterns of steady regional connectivity in the biological brain. It is noteworthy that during this process, the sparsity of long-range connections also gradually increases and stabilizes at around 50\%. This suggests that the process of applying acquired knowledge in the proposed algorithm selectively connects beneficial modules while avoiding connections with redundant modules, efficiently promoting positive knowledge transfer between tasks and enhancing the performance of progressively complex learning.
 
 Fig. \ref{f63}I (left) shows the long-range connection patterns between modules supported by the proposed algorithm. During learning, the evolutionary algorithm exhibits different selection probabilities for these patterns (Fig. \ref{f63}H and I, right). The most preferred connections integrate the output of the third module from a learned task into the input of the second module in a new task and the output of the third module into the input of the fourth module, facilitating interaction between later-stage modules. In addition, the second module of the previous task has fewer connections, and performance-driven evolutionary algorithms are more inclined to select underlying and deeper features for reuse in subsequent tasks. These results indicate that the long-range connections evolution and growth promote knowledge transfer, while local connection inhibition and pruning precisely remove redundant connections from learned tasks, allowing the proposed algorithm to retain acquired knowledge and support continual learning of new tasks during large-scale pruning.

\section{Discussion}

Inspired by the mechanism of temporal development of the brain, the proposed method breaks the traditional single-domain continual learning scenario and realizes perceptual-motor-interactive multiple cognitive task continual learning. Long-range connectivity evolutionary growth makes new tasks implemented based on old tasks outperform direct learning during forward knowledge transfer. Meanwhile progressive local pruning shrinks the network size as well as maintains the memory of the old task. Compared to the most widely used continual learning in visual perception, the proposed method demonstrates superior performance and stronger capability in learning complex tasks. 

Compared to existing continual learning algorithms, the proposed algorithm does not introduce common additional regularization loss terms\cite{kirkpatrick2017overcoming,chandra2023continual,batra2024evcl} or the replay of old task samples\cite{kurniawan2024evolving,jiang2025dupt} used in continual learning. At the same time, unlike structure-expansion continual learning algorithms\cite{han2023enhancing,han2025similarity}, TD-MCL prunes redundant connections in the structures of old tasks during new task learning, preserving the plasticity of old tasks without freezing their parameters. For biological plausibility, the proposed model aligns with the developmental sequence of biological brains from primary to advanced regions, as well as the developmental rule of gradually increasing inter-module communication\cite{wilke2007global} and initially increasing then decreasing intra-module connections\cite{huttenlocher1979synaptic}, providing an explainable method for positive knowledge transfer and memory capacity growth. 

In the proposed method, efficient multiple cognitive function continual learning relies on brain-inspired temporal growth and pruning, thus avoiding the dilemma where large-scale networks depend on parameter expansion to enhance cognitive abilities. Therefore, incorporating the brain's dynamic structural reorganization and compression mechanisms into large-scale networks may represent a promising direction for future research. Taking DeepSeek as an example, the optimization of computational efficiency through  Mixture-of-Experts (MoE) mechanisms provides a potential approach\cite{liu2024deepseek}. Building upon this, our future research aims to utilize brain-inspired, interpretable methods to advance large-scale models with improved energy efficiency, adaptability, and generalization capabilities.

\section{Method}

\subsection{Multiple cognitive Function Dataset Setup}

To simulate the continual enhancement of various cognitive abilities in the brain, we have designed a cross-domain perception-motion-interaction multiple cognitive function dataset. In the perception domain, we utilized three visual classification datasets, including sketches (3,929 images), cartoon drawings (3,929 images), and real photographs (1,670 images) \cite{zhou2020deep}, with image samples as inputs and recognition accuracy as the performance metric. For the motion domain, we employed the stand and walk tasks of the Walker agent and the Reacher agent task from the DeepMind Control Suite environment\cite{tassa2018deepmind} in Mujoco. Leveraging the image feature extraction capabilities learned from previous tasks, we also used images as inputs in motion tasks, learning new feature extraction while adaptively reusing the previously acquired basic feature extraction modules for color images. Each task was provided with 47,500 example videos and action output labels for supervised learning during training. During testing, actions were determined based on real-time environmental image inputs, with average return reward as the performance metric. In the interaction domain, building upon the acquired image and motion control capabilities, we used 10,000 example video images and state inputs from the drawer-open, hammer, and button-press tasks in the metaworld environment\cite{yu2020meta} during training. The outputs were continuous command sequences (real values). During testing, actions were similarly determined based on environmental images and states, with task success rate as the performance metric.

This comprehensive dataset and methodology aim to provide a robust framework for studying and enhancing multiple cognitive functions, facilitating advancements in artificial intelligence and cognitive science research.

\subsection{SNN Architecture Progressive Development}
Brain-inspired algorithms have garnered significant attention , among which Spiking Neural Networks (SNNs) provide a suitable foundational architecture. SNNs utilize brain-inspired spiking neurons as their basic units, transmitting information through discrete 0/1 spikes. As a result, SNNs exhibit high biological plausibility, energy efficiency, and strong adaptability to hardware platforms. In this paper, we employ the Parametric Leaky Integrate-and-Fire (PLIF) neuron \cite{fang2021incorporating}, with the membrane potential $U_i$ and spike $S_i$ as Eq. \ref{U}-\ref{S}, to construct a ResNet18 network architecture. The network weights are updated using a surrogate gradient algorithm, which enables effective training of the SNN while maintaining its spiking nature and computational efficiency. 

\begin{equation}
    U_i^{step}=\sigma(\tau) U_i^{step-1}+\sum_{j=1}^{M} P_t^{ij} S_j^{step}
    \label{U}
\end{equation}

\begin{equation}
	\label{S}
	S^{step}_{i}=\left\{\begin{matrix}
		1, &  U_i^{step}\geq V_{th}\\ 
		0, &  U_i^{step} < V_{th}
	\end{matrix}\right.
\end{equation}
where \(\tau\) is a learnable parameter, \(\sigma(\cdot)\) denotes the sigmoid function, and \(V_{th}\) represents the spike firing threshold.

During the first two years after birth, the infant brain continuously forms new connections to establish a structural foundation for progressively mastering diverse cognitive tasks \cite{sakai2020synaptic}. Inspired by this developmental mechanism, our algorithm similarly grows network modules to acquire previously unknown knowledge when learning new tasks $t$. Specifically, we expand the network horizontally at the block level for each convolutional layer, with the number of convolutional channels increasing by 32-64-128-256 across the four blocks of ResNet18 as Eq. \ref{T}. New modules are integrated via long-range connections to blocks associated with existing tasks, as detailed in the next section. 

\begin{equation}
	\label{T}
	\{B_{1}^{t},B_{2}^{t},B_{3}^{t},B_{4}^{t} \} \rightarrow \left\{\begin{matrix}
		B_{1}^{1},B_{2}^{1},B_{3}^{1},B_{4}^{1}\\ 
            \vdots \\
		B_{1}^{t-1},B_{2}^{t-1},B_{3}^{t-1},B_{4}^{t-1}
	\end{matrix}\right.
\end{equation}

\subsection{Adaptive evolution of long-range connectivity}
To avoid task interference caused by redundant connections and insufficient knowledge reuse due to missing connections, we do not manually design long-range connections between task modules. Instead, we adaptively learn the increasing inter-regional connections through an online evolutionary approach. Specifically, we evolve the connections between newly added modules in blocks 2-4 of ResNet18 and the existing groups of modules from previous tasks (block 1 receives the input). For each task $t$, the newly expanded module $b$ can choose whether to connect to the modules of previous tasks $k$ and which specific module to connect to. This selection is governed by a connection probability vector $P_{b}^{t,k}$ , as follow:

\begin{equation}
    P_{b}^{t,k}=\left[p_1,p_2,p_3,p_4,p_5,p_6\right]
    \label{P}
\end{equation}
Here, $p_1,p_2,p_3$ represent the probabilities of not connecting to any module from previous tasks $k$, resulting in approximately 50\% sparsity in long-range connections, consistent with biological data. Meanwhile, $p_4,p_5,p_6$ denote the probabilities of connecting to the three possible modules from task $k$. The connection probability matrix between task $t$ and task $k$ is $3
\cdot 6$. To maximize knowledge reuse, the connection probability matrix from expanded modules of task $t$ to all previous tasks is $3\cdot (t-1) \cdot 6$, ensuring efficient knowledge transfer while maintaining biological plausibility.

We update the connection probabilities based on the loss performance obtained from the current connection selections. First, we count the historical selection count $h_n$ and the loss obtained from selecting this connection $h_l=1-Normalize_{0-1}(loss)$. Then, we compare the selection counts and performance differences loss of each possible connection.

\begin{equation}
d_{h_n}= h_n - {h_n}^T , d_{h_l} = h_l - {h_l}^T
\label{dh}
\end{equation}
Where T denotes the transpose operation. Among these, potential connections that are used infrequently but exhibit high performance indicate their beneficial contribution to new tasks, and thus their connection probabilities are increased. Conversely, for potential connections that are used frequently but exhibit low performance, their selection probabilities are reduced. 
\begin{equation}
\begin{aligned}
dp^{+} &= \sum (d_{h_n}< 0 \land d_{h_l} > 0) \\
dp^{-} &= \sum (d_{h_n} > 0 \land d_{h_l} < 0) \
\end{aligned}
\label{dp}
\end{equation}

The learning rate for updating the connection probabilities $\gamma$ is set to 0.5, and the updated probabilities are normalized using the softmax function.
\begin{equation}
p= Softmax (p + \gamma \cdot (dp^{+} - dp^{-}) )
\label{pp}
\end{equation}

\subsection{Inhibition and pruning of local connections}

The pruning process in the developing brain occurs later than the initial learning period of infants, and instead acts on the continuous optimization of previous tasks\cite{loeffler2023neuromorphic}. Typically perceptual and motor brain regions prune early in the age window, whereas advanced brain regions such as the prefrontal cortex, which are responsible for interactive tasks, have a later age window\cite{huttenlocher1997regional}. Therefore, we provide feedback guidance to gradually inhibit and prune the local connectivity of learned task modules as we progressively learn new tasks. Specifically, inhibition and pruning are controlled by synaptic threshold coefficients that are related to local synaptic plasticity and global knowledge generalization of belonging modules. 

According to the principle of “use it or lose it”, we first calculate the hebbian synaptic plasticity, which composes of the spiking traces of presynaptic and postsynaptic neurons with the following formula:

\begin{equation}
Trace_i^{step}= \alpha Trace_i^{step}+ S_i^{step}
\label{T}
\end{equation}

\begin{equation}
H_{ij}=Normalize_{0-1}(Trace_{i}^{Step} \cdot {Trace_{j}^{Step}}^T)
\label{H}
\end{equation}
Where $\alpha$ is the decay coefficient. The $i$ and $j$ represent the post-synaptic and pre-synaptic neurons, respectively, while $step$ and $Step$ denote the spike timing $step$ and the total length of the spike time window, respectively.

In addition to the local spiking activity, we also consider the global knowledge generalization $E_b^k$ of the module across all tasks, represented by the sum of the probabilities that the module of the learned task $k$ will be used in subsequent tasks (up to the current task $t$), as shown in Eq. \ref{E}. 

\begin{equation}
E_b^k=P_b^{k+1,k}+P_b^{k+2,k}+....+P_b^{t,k}
\label{E}
\end{equation}
The high usage probability of these evolved modules indicates that their knowledge is generalizable and beneficial for subsequent tasks. In contrast, a low usage probability suggests the knowledge is specific to the current task and lacks generalizability. Pruning these redundant modules improves efficiency by removing non-essential connections.

In addition, brain pruning and inhibitory processes are gradually stabilized with age. Therefore, we combined local synaptic plasticity, global generalizability, and number of training runs to calculate the synaptic threshold coefficients, and ensured that the results were calculated between 0 and 1, as follow:

\begin{equation}
V_{ij}=min(1,n/N) \cdot (1 - e^{-2(H_{ij}+E_b^k)})
\label{V}
\end{equation}
Synaptic threshold coefficients were used to inhibit synapse weights or even prune synapses in varying degrees to improve network task focus and reduce overfitting, as shown in the following formula: 

\begin{equation}
w_{ij}^{'}=Sign(w_{ij})Relu(|w_{ij}|-V_{ij}|w_{ij}|_{\frac{4}{5}})
\label{w}
\end{equation}
where $w_{ij}$ and $w_{ij}^{'}$ denote original synaptic weights and synaptic weights after inhibition and pruning, respectively. $|w_{ij}|_{\frac{4}{5}}$ denotes the 0.8 quantile of the absolute value of synaptic weights.



\section*{Data availability}
The data used in this study are available in the following databases.

\noindent The perception cognitive task data~\cite{zhou2020deep}: \\ \href{https://github.com/robertofranceschi/Domain-adaptation-on-PACS-dataset}{https://github.com/robertofranceschi/Domain-adaptation-on-PACS-dataset}. 

\noindent The motion cognitive task DeepMind Control Suite environment data\cite{tassa2018deepmind}: \\ \href{https://dl.fbaipublicfiles.com/eai-vc/mujoco_vil_datasets/dmc-expert-v1.0.zip}{https://dl.fbaipublicfiles.com/eai-vc/mujoco$\_$vil$\_$datasets/dmc-expert-v1.0.zip}. 

\noindent The interaction cognitive task Metaworld environment data\cite{yu2020meta}: \\ \href{https://dl.fbaipublicfiles.com/eai-vc/mujoco_vil_datasets/metaworld-expert-v1.0.zip}{https://dl.fbaipublicfiles.com/eai-vc/mujoco$\_$vil$\_$datasets/metaworld-expert-v1.0.zip}.

\section*{Acknowledgments}
	This work is supported by the Strategic Priority Research Program of the Chinese Academy of Sciences (Grant No. XDB1010302), the National Natural Science Foundation of China (Grant No. 62106261), Chinese Academy of Sciences (Grant No. ZDBS-LY-JSC013). We are specially grateful to Dr. Mu-ming Poo for his invaluable guidance and inspiration. His profound academic insights and unwavering support were instrumental in the successful completion of this study.
 
\bibliographystyle{naturemag}
\bibliography{nature-communications}

\begin{thebibliography}{10}
\expandafter\ifx\csname url\endcsname\relax
  \def\url#1{\texttt{#1}}\fi
\expandafter\ifx\csname urlprefix\endcsname\relax\def\urlprefix{URL }\fi
\providecommand{\bibinfo}[2]{#2}
\providecommand{\eprint}[2][]{\url{#2}}

\bibitem{wickramasinghe2023continual}
\bibinfo{author}{Wickramasinghe, B.}, \bibinfo{author}{Saha, G.} \&
  \bibinfo{author}{Roy, K.}
\newblock \bibinfo{title}{Continual learning: A review of techniques,
  challenges, and future directions}.
\newblock \emph{\bibinfo{journal}{IEEE Transactions on Artificial
  Intelligence}} \textbf{\bibinfo{volume}{5}}, \bibinfo{pages}{2526--2546}
  (\bibinfo{year}{2023}).

\bibitem{huttenlocher1997regional}
\bibinfo{author}{Huttenlocher, P.~R.} \& \bibinfo{author}{Dabholkar, A.~S.}
\newblock \bibinfo{title}{Regional differences in synaptogenesis in human
  cerebral cortex}.
\newblock \emph{\bibinfo{journal}{Journal of comparative Neurology}}
  \textbf{\bibinfo{volume}{387}}, \bibinfo{pages}{167--178}
  (\bibinfo{year}{1997}).

\bibitem{johnson2001functional}
\bibinfo{author}{Johnson, M.~H.}
\newblock \bibinfo{title}{Functional brain development in humans}.
\newblock \emph{\bibinfo{journal}{Nature Reviews Neuroscience}}
  \textbf{\bibinfo{volume}{2}}, \bibinfo{pages}{475--483}
  (\bibinfo{year}{2001}).

\bibitem{huttenlocher1979synaptic}
\bibinfo{author}{Huttenlocher, P.~R.} \emph{et~al.}
\newblock \bibinfo{title}{Synaptic density in human frontal
  cortex-developmental changes and effects of aging}.
\newblock \emph{\bibinfo{journal}{Brain Res}} \textbf{\bibinfo{volume}{163}},
  \bibinfo{pages}{195--205} (\bibinfo{year}{1979}).

\bibitem{flavell1982cognitive}
\bibinfo{author}{Flavell, J.~H.}
\newblock \bibinfo{title}{On cognitive development}.
\newblock \emph{\bibinfo{journal}{Child development}} \bibinfo{pages}{1--10}
  (\bibinfo{year}{1982}).

\bibitem{zhang2021survey}
\bibinfo{author}{Zhang, Y.} \& \bibinfo{author}{Yang, Q.}
\newblock \bibinfo{title}{A survey on multi-task learning}.
\newblock \emph{\bibinfo{journal}{IEEE transactions on knowledge and data
  engineering}} \textbf{\bibinfo{volume}{34}}, \bibinfo{pages}{5586--5609}
  (\bibinfo{year}{2021}).

\bibitem{le2024continual}
\bibinfo{author}{Le, T.-T.}, \bibinfo{author}{Nguyen, M.},
  \bibinfo{author}{Nguyen, T.~T.}, \bibinfo{author}{Van, L.~N.} \&
  \bibinfo{author}{Nguyen, T.~H.}
\newblock \bibinfo{title}{Continual relation extraction via sequential
  multi-task learning}.
\newblock In \emph{\bibinfo{booktitle}{Proceedings of the AAAI Conference on
  Artificial Intelligence}}, vol.~\bibinfo{volume}{38},
  \bibinfo{pages}{18444--18452} (\bibinfo{year}{2024}).

\bibitem{marza2024task}
\bibinfo{author}{Marza, P.}, \bibinfo{author}{Matignon, L.},
  \bibinfo{author}{Simonin, O.} \& \bibinfo{author}{Wolf, C.}
\newblock \bibinfo{title}{Task-conditioned adaptation of visual features in
  multi-task policy learning}.
\newblock In \emph{\bibinfo{booktitle}{Proceedings of the IEEE/CVF Conference
  on Computer Vision and Pattern Recognition}}, \bibinfo{pages}{17847--17856}
  (\bibinfo{year}{2024}).

\bibitem{zhuang2020comprehensive}
\bibinfo{author}{Zhuang, F.} \emph{et~al.}
\newblock \bibinfo{title}{A comprehensive survey on transfer learning}.
\newblock \emph{\bibinfo{journal}{Proceedings of the IEEE}}
  \textbf{\bibinfo{volume}{109}}, \bibinfo{pages}{43--76}
  (\bibinfo{year}{2020}).

\bibitem{he2024efficient}
\bibinfo{author}{He, X.} \emph{et~al.}
\newblock \bibinfo{title}{An efficient knowledge transfer strategy for spiking
  neural networks from static to event domain}.
\newblock In \emph{\bibinfo{booktitle}{Proceedings of the AAAI Conference on
  Artificial Intelligence}}, vol.~\bibinfo{volume}{38},
  \bibinfo{pages}{512--520} (\bibinfo{year}{2024}).

\bibitem{zhao2024comparison}
\bibinfo{author}{Zhao, Z.}, \bibinfo{author}{Alzubaidi, L.},
  \bibinfo{author}{Zhang, J.}, \bibinfo{author}{Duan, Y.} \&
  \bibinfo{author}{Gu, Y.}
\newblock \bibinfo{title}{A comparison review of transfer learning and
  self-supervised learning: Definitions, applications, advantages and
  limitations}.
\newblock \emph{\bibinfo{journal}{Expert Systems with Applications}}
  \textbf{\bibinfo{volume}{242}}, \bibinfo{pages}{122807}
  (\bibinfo{year}{2024}).

\bibitem{van2019three}
\bibinfo{author}{Van~de Ven, G.~M.} \& \bibinfo{author}{Tolias, A.~S.}
\newblock \bibinfo{title}{Three scenarios for continual learning}.
\newblock \emph{\bibinfo{journal}{arXiv preprint arXiv:1904.07734}}
  (\bibinfo{year}{2019}).

\bibitem{wang2023task}
\bibinfo{author}{Wang, W.}, \bibinfo{author}{Hu, Y.}, \bibinfo{author}{Chen,
  Q.} \& \bibinfo{author}{Zhang, Y.}
\newblock \bibinfo{title}{Task difficulty aware parameter allocation \&
  regularization for lifelong learning}.
\newblock In \emph{\bibinfo{booktitle}{Proceedings of the IEEE/CVF Conference
  on Computer Vision and Pattern Recognition}}, \bibinfo{pages}{7776--7785}
  (\bibinfo{year}{2023}).

\bibitem{han2025similarity}
\bibinfo{author}{Han, B.} \emph{et~al.}
\newblock \bibinfo{title}{Similarity-based context aware continual learning for
  spiking neural networks}.
\newblock \emph{\bibinfo{journal}{Neural Networks}}
  \textbf{\bibinfo{volume}{184}}, \bibinfo{pages}{107037}
  (\bibinfo{year}{2025}).

\bibitem{hu2023dense}
\bibinfo{author}{Hu, Z.}, \bibinfo{author}{Li, Y.}, \bibinfo{author}{Lyu, J.},
  \bibinfo{author}{Gao, D.} \& \bibinfo{author}{Vasconcelos, N.}
\newblock \bibinfo{title}{Dense network expansion for class incremental
  learning}.
\newblock In \emph{\bibinfo{booktitle}{Proceedings of the IEEE/CVF Conference
  on Computer Vision and Pattern Recognition}}, \bibinfo{pages}{11858--11867}
  (\bibinfo{year}{2023}).

\bibitem{huttenlocher2009neural}
\bibinfo{author}{Huttenlocher, P.~R.}
\newblock \emph{\bibinfo{title}{Neural plasticity: The effects of environment
  on the development of the cerebral cortex}} (\bibinfo{publisher}{Harvard
  University Press}, \bibinfo{year}{2009}).

\bibitem{hagmann2010white}
\bibinfo{author}{Hagmann, P.} \emph{et~al.}
\newblock \bibinfo{title}{White matter maturation reshapes structural
  connectivity in the late developing human brain}.
\newblock \emph{\bibinfo{journal}{Proceedings of the National Academy of
  Sciences}} \textbf{\bibinfo{volume}{107}}, \bibinfo{pages}{19067--19072}
  (\bibinfo{year}{2010}).

\bibitem{selemon2013role}
\bibinfo{author}{Selemon, L.~D.}
\newblock \bibinfo{title}{A role for synaptic plasticity in the adolescent
  development of executive function}.
\newblock \emph{\bibinfo{journal}{Translational psychiatry}}
  \textbf{\bibinfo{volume}{3}}, \bibinfo{pages}{e238--e238}
  (\bibinfo{year}{2013}).

\bibitem{scott2004optimal}
\bibinfo{author}{Scott, S.~H.}
\newblock \bibinfo{title}{Optimal feedback control and the neural basis of
  volitional motor control}.
\newblock \emph{\bibinfo{journal}{Nature Reviews Neuroscience}}
  \textbf{\bibinfo{volume}{5}}, \bibinfo{pages}{532--545}
  (\bibinfo{year}{2004}).

\bibitem{sitaram2017closed}
\bibinfo{author}{Sitaram, R.} \emph{et~al.}
\newblock \bibinfo{title}{Closed-loop brain training: the science of
  neurofeedback}.
\newblock \emph{\bibinfo{journal}{Nature Reviews Neuroscience}}
  \textbf{\bibinfo{volume}{18}}, \bibinfo{pages}{86--100}
  (\bibinfo{year}{2017}).

\bibitem{finlay2001developmental}
\bibinfo{author}{Finlay, B.~L.}, \bibinfo{author}{Darlington, R.~B.} \&
  \bibinfo{author}{Nicastro, N.}
\newblock \bibinfo{title}{Developmental structure in brain evolution}.
\newblock \emph{\bibinfo{journal}{Behavioral and Brain Sciences}}
  \textbf{\bibinfo{volume}{24}}, \bibinfo{pages}{263--278}
  (\bibinfo{year}{2001}).

\bibitem{shi2023towards}
\bibinfo{author}{Shi, X.}, \bibinfo{author}{Ding, J.}, \bibinfo{author}{Hao,
  Z.} \& \bibinfo{author}{Yu, Z.}
\newblock \bibinfo{title}{Towards energy efficient spiking neural networks: An
  unstructured pruning framework}.
\newblock In \emph{\bibinfo{booktitle}{The Twelfth International Conference on
  Learning Representations}} (\bibinfo{year}{2023}).

\bibitem{yu2022width}
\bibinfo{author}{Yu, F.} \emph{et~al.}
\newblock \bibinfo{title}{Width \& depth pruning for vision transformers}.
\newblock In \emph{\bibinfo{booktitle}{Proceedings of the AAAI Conference on
  Artificial Intelligence}}, vol.~\bibinfo{volume}{36},
  \bibinfo{pages}{3143--3151} (\bibinfo{year}{2022}).

\bibitem{balaskas2024hardware}
\bibinfo{author}{Balaskas, K.} \emph{et~al.}
\newblock \bibinfo{title}{Hardware-aware dnn compression via diverse pruning
  and mixed-precision quantization}.
\newblock \emph{\bibinfo{journal}{IEEE Transactions on Emerging Topics in
  Computing}} \textbf{\bibinfo{volume}{12}}, \bibinfo{pages}{1079--1092}
  (\bibinfo{year}{2024}).

\bibitem{molchanov2016pruning}
\bibinfo{author}{Molchanov, P.}, \bibinfo{author}{Tyree, S.},
  \bibinfo{author}{Karras, T.}, \bibinfo{author}{Aila, T.} \&
  \bibinfo{author}{Kautz, J.}
\newblock \bibinfo{title}{Pruning convolutional neural networks for resource
  efficient inference}.
\newblock \emph{\bibinfo{journal}{arXiv preprint arXiv:1611.06440}}
  (\bibinfo{year}{2016}).

\bibitem{you2019gate}
\bibinfo{author}{You, Z.}, \bibinfo{author}{Yan, K.}, \bibinfo{author}{Ye, J.},
  \bibinfo{author}{Ma, M.} \& \bibinfo{author}{Wang, P.}
\newblock \bibinfo{title}{Gate decorator: Global filter pruning method for
  accelerating deep convolutional neural networks}.
\newblock \emph{\bibinfo{journal}{Advances in neural information processing
  systems}} \textbf{\bibinfo{volume}{32}} (\bibinfo{year}{2019}).

\bibitem{srinivas2015data}
\bibinfo{author}{Srinivas, S.} \& \bibinfo{author}{Babu, R.~V.}
\newblock \bibinfo{title}{Data-free parameter pruning for deep neural
  networks}.
\newblock \emph{\bibinfo{journal}{arXiv preprint arXiv:1507.06149}}
  (\bibinfo{year}{2015}).

\bibitem{Maass1997Networks}
\bibinfo{author}{Maass, W.}
\newblock \bibinfo{title}{Networks of spiking neurons: The third generation of
  neural network models}.
\newblock \emph{\bibinfo{journal}{Neural networks}}
  \textbf{\bibinfo{volume}{10}}, \bibinfo{pages}{1659--1671}
  (\bibinfo{year}{1997}).

\bibitem{gerstner2002spiking}
\bibinfo{author}{Gerstner, W.} \& \bibinfo{author}{Kistler, W.~M.}
\newblock \emph{\bibinfo{title}{Spiking neuron models: Single neurons,
  populations, plasticity}} (\bibinfo{publisher}{Cambridge university press},
  \bibinfo{year}{2002}).

\bibitem{rathi2018stdp}
\bibinfo{author}{Rathi, N.}, \bibinfo{author}{Panda, P.} \&
  \bibinfo{author}{Roy, K.}
\newblock \bibinfo{title}{Stdp-based pruning of connections and weight
  quantization in spiking neural networks for energy-efficient recognition}.
\newblock \emph{\bibinfo{journal}{IEEE Transactions on Computer-Aided Design of
  Integrated Circuits and Systems}} \textbf{\bibinfo{volume}{38}},
  \bibinfo{pages}{668--677} (\bibinfo{year}{2018}).

\bibitem{han2024developmental}
\bibinfo{author}{Han, B.}, \bibinfo{author}{Zhao, F.}, \bibinfo{author}{Zeng,
  Y.} \& \bibinfo{author}{Shen, G.}
\newblock \bibinfo{title}{Developmental plasticity-inspired adaptive pruning
  for deep spiking and artificial neural networks}.
\newblock \emph{\bibinfo{journal}{IEEE Transactions on Pattern Analysis and
  Machine Intelligence}}  (\bibinfo{year}{2024}).

\bibitem{wu2019adaptive}
\bibinfo{author}{Wu, D.}, \bibinfo{author}{Lin, X.} \& \bibinfo{author}{Du, P.}
\newblock \bibinfo{title}{An adaptive structure learning algorithm for
  multi-layer spiking neural networks}.
\newblock In \emph{\bibinfo{booktitle}{2019 15th International Conference on
  Computational Intelligence and Security (CIS)}}, \bibinfo{pages}{98--102}
  (\bibinfo{organization}{IEEE}, \bibinfo{year}{2019}).

\bibitem{liu2022dynsnn}
\bibinfo{author}{Liu, F.}, \bibinfo{author}{Zhao, W.}, \bibinfo{author}{Chen,
  Y.}, \bibinfo{author}{Wang, Z.} \& \bibinfo{author}{Dai, F.}
\newblock \bibinfo{title}{Dynsnn: A dynamic approach to reduce redundancy in
  spiking neural networks}.
\newblock In \emph{\bibinfo{booktitle}{ICASSP 2022-2022 IEEE International
  Conference on Acoustics, Speech and Signal Processing (ICASSP)}},
  \bibinfo{pages}{2130--2134} (\bibinfo{organization}{IEEE},
  \bibinfo{year}{2022}).

\bibitem{chen2022state}
\bibinfo{author}{Chen, Y.} \emph{et~al.}
\newblock \bibinfo{title}{State transition of dendritic spines improves
  learning of sparse spiking neural networks}.
\newblock In \emph{\bibinfo{booktitle}{International Conference on Machine
  Learning}}, \bibinfo{pages}{3701--3715} (\bibinfo{organization}{PMLR},
  \bibinfo{year}{2022}).

\bibitem{han2025adaptive}
\bibinfo{author}{Han, B.}, \bibinfo{author}{Zhao, F.}, \bibinfo{author}{Pan,
  W.} \& \bibinfo{author}{Zeng, Y.}
\newblock \bibinfo{title}{Adaptive sparse structure development with pruning
  and regeneration for spiking neural networks}.
\newblock \emph{\bibinfo{journal}{Information Sciences}}
  \textbf{\bibinfo{volume}{689}}, \bibinfo{pages}{121481}
  (\bibinfo{year}{2025}).

\bibitem{inguaggiato2017brain}
\bibinfo{author}{Inguaggiato, E.}, \bibinfo{author}{Sgandurra, G.} \&
  \bibinfo{author}{Cioni, G.}
\newblock \bibinfo{title}{Brain plasticity and early development: Implications
  for early intervention in neurodevelopmental disorders}.
\newblock \emph{\bibinfo{journal}{Neuropsychiatrie de l'Enfance et de
  l'Adolescence}} \textbf{\bibinfo{volume}{65}}, \bibinfo{pages}{299--306}
  (\bibinfo{year}{2017}).

\bibitem{zhou2020deep}
\bibinfo{author}{Zhou, K.}, \bibinfo{author}{Yang, Y.},
  \bibinfo{author}{Hospedales, T.} \& \bibinfo{author}{Xiang, T.}
\newblock \bibinfo{title}{Deep domain-adversarial image generation for domain
  generalisation}.
\newblock In \emph{\bibinfo{booktitle}{Proceedings of the AAAI conference on
  artificial intelligence}}, vol.~\bibinfo{volume}{34},
  \bibinfo{pages}{13025--13032} (\bibinfo{year}{2020}).

\bibitem{tassa2018deepmind}
\bibinfo{author}{Tassa, Y.} \emph{et~al.}
\newblock \bibinfo{title}{Deepmind control suite}.
\newblock \emph{\bibinfo{journal}{arXiv preprint arXiv:1801.00690}}
  (\bibinfo{year}{2018}).

\bibitem{yu2020meta}
\bibinfo{author}{Yu, T.} \emph{et~al.}
\newblock \bibinfo{title}{Meta-world: A benchmark and evaluation for multi-task
  and meta reinforcement learning}.
\newblock In \emph{\bibinfo{booktitle}{Conference on robot learning}},
  \bibinfo{pages}{1094--1100} (\bibinfo{organization}{PMLR},
  \bibinfo{year}{2020}).

\bibitem{huttenlocher2013synaptogenesis}
\bibinfo{author}{Huttenlocher, P.~R.}
\newblock \bibinfo{title}{Synaptogenesis, synapse elimination, and neural
  plasticity in human cerebral cortex}.
\newblock In \emph{\bibinfo{booktitle}{Threats to optimal development}},
  \bibinfo{pages}{35--54} (\bibinfo{publisher}{Routledge},
  \bibinfo{year}{2013}).

\bibitem{bourgeois1993changes}
\bibinfo{author}{Bourgeois, J.-P.} \& \bibinfo{author}{Rakic, P.}
\newblock \bibinfo{title}{Changes of synaptic density in the primary visual
  cortex of the macaque monkey from fetal to adult stage}.
\newblock \emph{\bibinfo{journal}{Journal of Neuroscience}}
  \textbf{\bibinfo{volume}{13}}, \bibinfo{pages}{2801--2820}
  (\bibinfo{year}{1993}).

\bibitem{kroon2019early}
\bibinfo{author}{Kroon, T.}, \bibinfo{author}{van Hugte, E.},
  \bibinfo{author}{van Linge, L.}, \bibinfo{author}{Mansvelder, H.~D.} \&
  \bibinfo{author}{Meredith, R.~M.}
\newblock \bibinfo{title}{Early postnatal development of pyramidal neurons
  across layers of the mouse medial prefrontal cortex}.
\newblock \emph{\bibinfo{journal}{Scientific reports}}
  \textbf{\bibinfo{volume}{9}}, \bibinfo{pages}{5037} (\bibinfo{year}{2019}).

\bibitem{wilke2007global}
\bibinfo{author}{Wilke, M.}, \bibinfo{author}{Kr{\"a}geloh-Mann, I.} \&
  \bibinfo{author}{Holland, S.~K.}
\newblock \bibinfo{title}{Global and local development of gray and white matter
  volume in normal children and adolescents}.
\newblock \emph{\bibinfo{journal}{Experimental Brain Research}}
  \textbf{\bibinfo{volume}{178}}, \bibinfo{pages}{296--307}
  (\bibinfo{year}{2007}).

\bibitem{groeschel2010developmental}
\bibinfo{author}{Groeschel, S.}, \bibinfo{author}{Vollmer, B.},
  \bibinfo{author}{King, M.} \& \bibinfo{author}{Connelly, A.}
\newblock \bibinfo{title}{Developmental changes in cerebral grey and white
  matter volume from infancy to adulthood}.
\newblock \emph{\bibinfo{journal}{International Journal of Developmental
  Neuroscience}} \textbf{\bibinfo{volume}{28}}, \bibinfo{pages}{481--489}
  (\bibinfo{year}{2010}).

\bibitem{giedd2015child}
\bibinfo{author}{Giedd, J.~N.} \emph{et~al.}
\newblock \bibinfo{title}{Child psychiatry branch of the national institute of
  mental health longitudinal structural magnetic resonance imaging study of
  human brain development}.
\newblock \emph{\bibinfo{journal}{Neuropsychopharmacology}}
  \textbf{\bibinfo{volume}{40}}, \bibinfo{pages}{43--49}
  (\bibinfo{year}{2015}).

\bibitem{reveley2015superficial}
\bibinfo{author}{Reveley, C.} \emph{et~al.}
\newblock \bibinfo{title}{Superficial white matter fiber systems impede
  detection of long-range cortical connections in diffusion mr tractography}.
\newblock \emph{\bibinfo{journal}{Proceedings of the National Academy of
  Sciences}} \textbf{\bibinfo{volume}{112}}, \bibinfo{pages}{E2820--E2828}
  (\bibinfo{year}{2015}).

\bibitem{schuz2002human}
\bibinfo{author}{Sch{\"u}z, A.} \& \bibinfo{author}{Braitenberg, V.}
\newblock \bibinfo{title}{The human cortical white matter: quantitative aspects
  of cortico-cortical long-range connectivity}.
\newblock In \emph{\bibinfo{booktitle}{Cortical areas}},
  \bibinfo{pages}{389--398} (\bibinfo{publisher}{CRC Press},
  \bibinfo{year}{2002}).

\bibitem{chechik1999neuronal}
\bibinfo{author}{Chechik, G.}, \bibinfo{author}{Meilijson, I.} \&
  \bibinfo{author}{Ruppin, E.}
\newblock \bibinfo{title}{Neuronal regulation: A mechanism for synaptic pruning
  during brain maturation}.
\newblock \emph{\bibinfo{journal}{Neural computation}}
  \textbf{\bibinfo{volume}{11}}, \bibinfo{pages}{2061--2080}
  (\bibinfo{year}{1999}).

\bibitem{stephan2012complement}
\bibinfo{author}{Stephan, A.~H.}, \bibinfo{author}{Barres, B.~A.} \&
  \bibinfo{author}{Stevens, B.}
\newblock \bibinfo{title}{The complement system: an unexpected role in synaptic
  pruning during development and disease}.
\newblock \emph{\bibinfo{journal}{Annual review of neuroscience}}
  \textbf{\bibinfo{volume}{35}}, \bibinfo{pages}{369--389}
  (\bibinfo{year}{2012}).

\bibitem{faust2021mechanisms}
\bibinfo{author}{Faust, T.~E.}, \bibinfo{author}{Gunner, G.} \&
  \bibinfo{author}{Schafer, D.~P.}
\newblock \bibinfo{title}{Mechanisms governing activity-dependent synaptic
  pruning in the developing mammalian cns}.
\newblock \emph{\bibinfo{journal}{Nature Reviews Neuroscience}}
  \textbf{\bibinfo{volume}{22}}, \bibinfo{pages}{657--673}
  (\bibinfo{year}{2021}).

\bibitem{ugarte2023attention}
\bibinfo{author}{Ugarte, G.} \emph{et~al.}
\newblock \bibinfo{title}{Attention deficit-hyperactivity disorder (adhd): from
  abnormal behavior to impairment in synaptic plasticity}.
\newblock \emph{\bibinfo{journal}{Biology}} \textbf{\bibinfo{volume}{12}},
  \bibinfo{pages}{1241} (\bibinfo{year}{2023}).

\bibitem{kirkpatrick2017overcoming}
\bibinfo{author}{Kirkpatrick, J.} \emph{et~al.}
\newblock \bibinfo{title}{Overcoming catastrophic forgetting in neural
  networks}.
\newblock \emph{\bibinfo{journal}{Proceedings of the national academy of
  sciences}} \textbf{\bibinfo{volume}{114}}, \bibinfo{pages}{3521--3526}
  (\bibinfo{year}{2017}).

\bibitem{aljundi2018memory}
\bibinfo{author}{Aljundi, R.}, \bibinfo{author}{Babiloni, F.},
  \bibinfo{author}{Elhoseiny, M.}, \bibinfo{author}{Rohrbach, M.} \&
  \bibinfo{author}{Tuytelaars, T.}
\newblock \bibinfo{title}{Memory aware synapses: Learning what (not) to
  forget}.
\newblock In \emph{\bibinfo{booktitle}{Proceedings of the European Conference
  on Computer Vision (ECCV)}}, \bibinfo{pages}{139--154}
  (\bibinfo{year}{2018}).

\bibitem{von2019continual}
\bibinfo{author}{Von~Oswald, J.}, \bibinfo{author}{Henning, C.},
  \bibinfo{author}{Grewe, B.~F.} \& \bibinfo{author}{Sacramento, J.}
\newblock \bibinfo{title}{Continual learning with hypernetworks}.
\newblock \emph{\bibinfo{journal}{arXiv preprint arXiv:1906.00695}}
  (\bibinfo{year}{2019}).

\bibitem{chandra2023continual}
\bibinfo{author}{Chandra, D.~S.}, \bibinfo{author}{Varshney, S.},
  \bibinfo{author}{Srijith, P.} \& \bibinfo{author}{Gupta, S.}
\newblock \bibinfo{title}{Continual learning with dependency preserving
  hypernetworks}.
\newblock In \emph{\bibinfo{booktitle}{Proceedings of the IEEE/CVF Winter
  Conference on Applications of Computer Vision}}, \bibinfo{pages}{2339--2348}
  (\bibinfo{year}{2023}).

\bibitem{pham2021dualnet}
\bibinfo{author}{Pham, Q.}, \bibinfo{author}{Liu, C.} \& \bibinfo{author}{Hoi,
  S.}
\newblock \bibinfo{title}{Dualnet: Continual learning, fast and slow}.
\newblock \emph{\bibinfo{journal}{Advances in Neural Information Processing
  Systems}} \textbf{\bibinfo{volume}{34}}, \bibinfo{pages}{16131--16144}
  (\bibinfo{year}{2021}).

\bibitem{han2023enhancing}
\bibinfo{author}{Han, B.}, \bibinfo{author}{Zhao, F.}, \bibinfo{author}{Zeng,
  Y.}, \bibinfo{author}{Pan, W.} \& \bibinfo{author}{Shen, G.}
\newblock \bibinfo{title}{Enhancing efficient continual learning with dynamic
  structure development of spiking neural networks}.
\newblock \emph{\bibinfo{journal}{arXiv preprint arXiv:2308.04749}}
  (\bibinfo{year}{2023}).

\bibitem{han2023adaptive}
\bibinfo{author}{Han, B.} \emph{et~al.}
\newblock \bibinfo{title}{Adaptive reorganization of neural pathways for
  continual learning with spiking neural networks}.
\newblock \emph{\bibinfo{journal}{arXiv preprint arXiv:2309.09550}}
  (\bibinfo{year}{2023}).

\bibitem{favazza2016motor}
\bibinfo{author}{Favazza, P.~C.} \& \bibinfo{author}{Siperstein, G.~N.}
\newblock \bibinfo{title}{Motor skill acquisition for young children with
  disabilities}.
\newblock \emph{\bibinfo{journal}{Handbook of early childhood special
  education}} \bibinfo{pages}{225--245} (\bibinfo{year}{2016}).

\bibitem{batra2024evcl}
\bibinfo{author}{Batra, H.} \& \bibinfo{author}{Clark, R.}
\newblock \bibinfo{title}{Evcl: Elastic variational continual learning with
  weight consolidation}.
\newblock In \emph{\bibinfo{booktitle}{ICML 2024 Workshop on Structured
  Probabilistic Inference $\{$$\backslash$\&$\}$ Generative Modeling}}.

\bibitem{kurniawan2024evolving}
\bibinfo{author}{Kurniawan, M.~R.} \emph{et~al.}
\newblock \bibinfo{title}{Evolving parameterized prompt memory for continual
  learning}.
\newblock In \emph{\bibinfo{booktitle}{Proceedings of the AAAI Conference on
  Artificial Intelligence}}, vol.~\bibinfo{volume}{38},
  \bibinfo{pages}{13301--13309} (\bibinfo{year}{2024}).

\bibitem{jiang2025dupt}
\bibinfo{author}{Jiang, S.}, \bibinfo{author}{Zhang, D.},
  \bibinfo{author}{Cheng, F.}, \bibinfo{author}{Lu, X.} \&
  \bibinfo{author}{Liu, Q.}
\newblock \bibinfo{title}{Dupt: Rehearsal-based continual learning with dual
  prompts}.
\newblock \emph{\bibinfo{journal}{Neural Networks}} \bibinfo{pages}{107306}
  (\bibinfo{year}{2025}).

\bibitem{liu2024deepseek}
\bibinfo{author}{Liu, A.} \emph{et~al.}
\newblock \bibinfo{title}{Deepseek-v3 technical report}.
\newblock \emph{\bibinfo{journal}{arXiv preprint arXiv:2412.19437}}
  (\bibinfo{year}{2024}).

\bibitem{fang2021incorporating}
\bibinfo{author}{Fang, W.} \emph{et~al.}
\newblock \bibinfo{title}{Incorporating learnable membrane time constant to
  enhance learning of spiking neural networks}.
\newblock In \emph{\bibinfo{booktitle}{Proceedings of the IEEE/CVF
  international conference on computer vision}}, \bibinfo{pages}{2661--2671}
  (\bibinfo{year}{2021}).

\bibitem{sakai2020synaptic}
\bibinfo{author}{Sakai, J.}
\newblock \bibinfo{title}{How synaptic pruning shapes neural wiring during
  development and, possibly, in disease}.
\newblock \emph{\bibinfo{journal}{Proceedings of the National Academy of
  Sciences}} \textbf{\bibinfo{volume}{117}}, \bibinfo{pages}{16096--16099}
  (\bibinfo{year}{2020}).

\bibitem{loeffler2023neuromorphic}
\bibinfo{author}{Loeffler, A.} \emph{et~al.}
\newblock \bibinfo{title}{Neuromorphic learning, working memory, and
  metaplasticity in nanowire networks}.
\newblock \emph{\bibinfo{journal}{Science Advances}}
  \textbf{\bibinfo{volume}{9}}, \bibinfo{pages}{eadg3289}
  (\bibinfo{year}{2023}).

\end{thebibliography}

\section*{Contributions}
B.H.,F.Z. and Y.Z. designed the study. B.H.,F.Z. Y.S. and W.P.performed the experiments. B.H., F.Z. and Y.S. counted and analyzed the experiment results. B.H.,F.Z., Y.S. and Y.Z. wrote the paper.

\section*{Competing interests}
The authors declare no competing interests.

\end{document}